\documentclass[letterpaper]{article} 
\usepackage{aaai2026}  
\usepackage{times}  
\usepackage{helvet}  
\usepackage{courier}  
\usepackage[hyphens]{url}  
\usepackage{graphicx} 
\usepackage{natbib}  
\usepackage{caption} 
\usepackage{algorithm}
\usepackage{algorithmic}

\usepackage{newfloat}
\usepackage{listings}
\DeclareCaptionStyle{ruled}{labelfont=normalfont,labelsep=colon,strut=off} 
\floatstyle{ruled}
\newfloat{listing}{tb}{lst}{}
\floatname{listing}{Listing}
\usepackage[utf8]{inputenc} 
\usepackage{url}            
\usepackage{booktabs}       
\usepackage[table]{xcolor} 
\usepackage{amsfonts}       
\usepackage{amsmath}        
\usepackage{nicefrac}       
\usepackage{microtype}      
\usepackage{xcolor}         
\usepackage{booktabs}
\usepackage{multirow} 
\usepackage{array}   

\newcommand{\mn}{LAid}

\title{Towards Long-window Anchoring in Vision-Language Model Distillation}
\author{
    Haoyi Zhou\textsuperscript{\rm 1},
    Shuo Li\textsuperscript{\rm 2},
    Tianyu Chen\textsuperscript{\rm 2},
    Qi Song\textsuperscript{\rm 1},
    Chonghan Gao\textsuperscript{\rm 2},
    Jianxin Li\textsuperscript{\rm 2, \rm 3}\thanks{Corresponding Author.}
}
\affiliations{
    \textsuperscript{\rm 1}School of Software, Beihang University, Beijing, China\\
    \textsuperscript{\rm 2}SKLCCSE, School of Computer Science and Engineering, Beihang University, Beijing , China\\
    \textsuperscript{\rm 3}Zhongguancun Laboratory, Beijing, China

    \{haoyi, lishuo2001, tianyuc, songqi23, gaoch, lijx\}@buaa.edu.cn
}

\begin{document}

\maketitle

\begin{abstract}
  While large vision-language models (VLMs) demonstrate strong long-context understanding, their prevalent small branches fail on linguistics-photography alignment for a limited window size. We discover that knowledge distillation improves students' capability as a complement to Rotary Position Embeddings (RoPE) on window sizes (anchored from large models). Building on this insight, we propose \mn, which directly aims at the transfer of long-range attention mechanisms through two complementary components: (1) a progressive distance-weighted attention matching that dynamically emphasizes longer position differences during training, and (2) a learnable RoPE response gain modulation that selectively amplifies position sensitivity where needed. Extensive experiments across multiple model families demonstrate that \mn-distilled models achieve up to 3.2× longer effective context windows compared to baseline small models, while maintaining or improving performance on standard VL benchmarks. Spectral analysis also suggests that \mn~successfully preserves crucial low-frequency attention components that conventional methods fail to transfer. Our work not only provides practical techniques for building more efficient long-context VLMs but also offers theoretical insights into how positional understanding emerges and transfers during distillation.

\end{abstract}


\section{Introduction}

The comprehensive understanding and full utilization of long context play a crucial role in building large vision-language models (VLMs). It brings better linguistics-photography alignment in both large scene~\cite{Claude3,qwen25vl, Gemma3, Llama4,LongViLa,internvl3} and long storyline~\cite{qwen25vl, Llama4,longllama,Yu2024long,internvl3}, and it helps improving coherence and depth of interaction in multi-rounds dialogue~\cite{Claude3,qwen25vl,Ge2024Discard,Gemma3,Llama4,longllama,llava-di,Yi-01,internvl3}. Currently, large-scale VLMs ($\geq$ 72B parameters) demonstrate the window size scales up to 128k tokens, e.g., Gemma 3~\cite{Gemma3}, Qwen 2.5-VL~\cite{qwen25vl}, InternVL 3~\cite{internvl3}. To the best of our knowledge, we are the first to point out that the VLMs' prevalent distilled branches ($\leq$ 7B parameters) exhibit markedly constrained window size despite their employment of exact positional embedding, identical architecture, and training methodology. This window shrink is negligible when applying distilled models on short context evaluations, but it becomes a major obstacle during full-length inferencing. 

\begin{figure}[tbp]
    \centering
    \includegraphics[width=\linewidth]{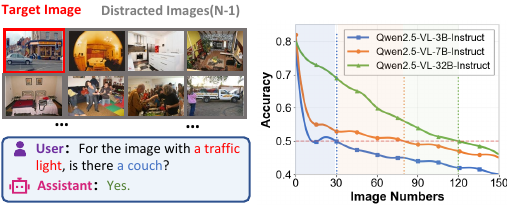}
   \caption{Effective context window comparison. \textbf{Left:} The Visual Haystack task requiring retrieval from multi-image inputs. \textbf{Right:} Qwen2.5-VL accuracy across scales. Larger models (32B) sustain effective performance ($>$0.5) significantly longer than smaller counterparts (3B, 7B) despite identical architectures, revealing a scale-dependent RoPE awareness gap that our method targets.}
    \label{fig:Motivation}
\end{figure}

Previous studies have revealed the possibility of extending the pre-trained large language models (LLMs) context window to more than 10 million tokens through various training-stage interventions. Position embedding extrapolation techniques become predominant, e.g., RoPE~\cite{Su2024Roformer} enabling models to generalize to sequence lengths beyond their training range, ALiBi~\cite{alibi} introducing inductive biases that scale effectively to longer contexts, LongRoPE~\cite{longRoPE} and FoPE~\cite{FoPE} leveraging the non-uniformity positional interpolation. Besides, fine-tuning approaches on longer texts have shown remarkable effectiveness, as demonstrated in works like LongLLaMA~\cite{longllama}, which extended context windows through continued pre-training on carefully curated long documents, and Anthropic's Claude model~\cite{Claude3}, which achieved 100k token context through specialized training regimes.
The above methods mainly focus on the training stage, requiring significant computational resources for model retraining or fine-tuning. 

However, VLMs encounter unique hurdles in handling long windows during training due to their multimodal nature—the visual components introduce substantial complexity in positional understanding across modalities, memory constraints become more severe due to image token density, and the alignment between visual and textual elements at long distances requires fundamentally different mechanisms. More importantly, many foundational assumptions that underpin text-only context extension techniques -- such as uniform attention patterns and sequence-independent position encodings -- break down when visual tokens with dense, spatially-organized information are introduced into the context window. Few research efforts have thrown light on post-training stage techniques that could extend context windows without expensive retraining.

To better motivate our work, we specialize the concept of \textbf{\textit{Long-window Anchoring}}. Current state-of-the-art VLMs (Qwen2.5-VL~\cite{qwen25vl}, InternVL 3~\cite{internvl3}, Gemma~\cite{Gemma3}) often develop models of varying parameter sizes(3B, 7B, 32B) through independent training from scratch, resulting in inconsistent window size capabilities across model sizes. We propose using larger models (e.g., Qwen 2.5-VL 32B) as ``anchors'' that possess strong long-window capability, then employing post-training methods to align smaller models' long-window capability with these anchors. In this way, smaller models can inherit long-window capability without the prohibitive computational cost of training from scratch, while maintaining their efficiency advantages. 
Among various potential post-training methods, we begin with knowledge distillation as our fundamental approach -- a proven technique for transferring capabilities between models of different sizes while maintaining computational efficiency for the smaller target model.


In this paper, we propose a new perspective to analyze this phenomenon in Figure \ref{fig:Motivation}, where we uncover a fundamental distinction: larger VLMs inherently sustain stronger visual haystack performance at extended input image numbers, decaying 5.2× slower compared to 3B models. This positional awareness gap persists even when smaller models demonstrate near-perfect performance on short-context tasks, suggesting the difference stems not from general capacity limitations but specifically from positional representation capabilities. Our analysis reveals that while standard knowledge distillation can unintentionally enhance students' RoPE responsiveness, this effect remains suboptimal without explicit long-context optimization.

To address this challenge, we propose \underline{L}ong-window \underline{A}nchor\underline{i}ng \underline{d}istillation (LAid), a distillation framework that explicitly targets the transfer of long-range attention mechanisms. LAid leverages a Fourier perspective on position distillation through head-level alignment, where each student head learns a weighted combination of multiple teacher heads' query and key representations: $Q^s_{l,i} \approx \sum^{ht}_{j=1} w_{i,j} \cdot Q^t_{L,j}, \ K^s_{l,i} \approx \sum^{ht}_{j=1} w_{i,j} \cdot K^t_{L,j}$. This formulation enables the student to acquire enhanced rotational encoding: $R'\theta(m) = \sum^{ht}_{j=1} w_{i,j} \cdot (W^Q_{t,j} \cdot R_\theta(m) \cdot (W^Q_{t,j})^{-1})$, which expands beyond the frequency limitations of standard RoPE and mitigates frequency leakage in smaller models. Our complete distillation objective combines this position-aware component with traditional knowledge distillation losses, creating a balanced approach that preserves both task performance and positional understanding across contexts of varying lengths.

Our contributions are threefold:
(1)  We formulate Long-window Anchoring as a distinct problem from traditional context extension—focusing on elevating small models' effective window lengths toward teacher models' upper bounds while preserving computational efficiency; 
(2)  We introduce LAid, a novel distillation framework leveraging head-level alignment with Fourier-enhanced positional knowledge transfer to overcome frequency leakage in compact architectures;
(3)  We empirically demonstrate the unique challenges vision-language models face with extended contexts—where even 32B-parameter models achieve merely 62.56\% accuracy at 100 images, substantially underperforming text-only counterparts. Our approach bridges this capability gap, extending effective context windows by up to 3.2× while preserving overall performance on standard benchmarks.

\begin{figure*}[t]
    \centering
    \includegraphics[width=\linewidth]{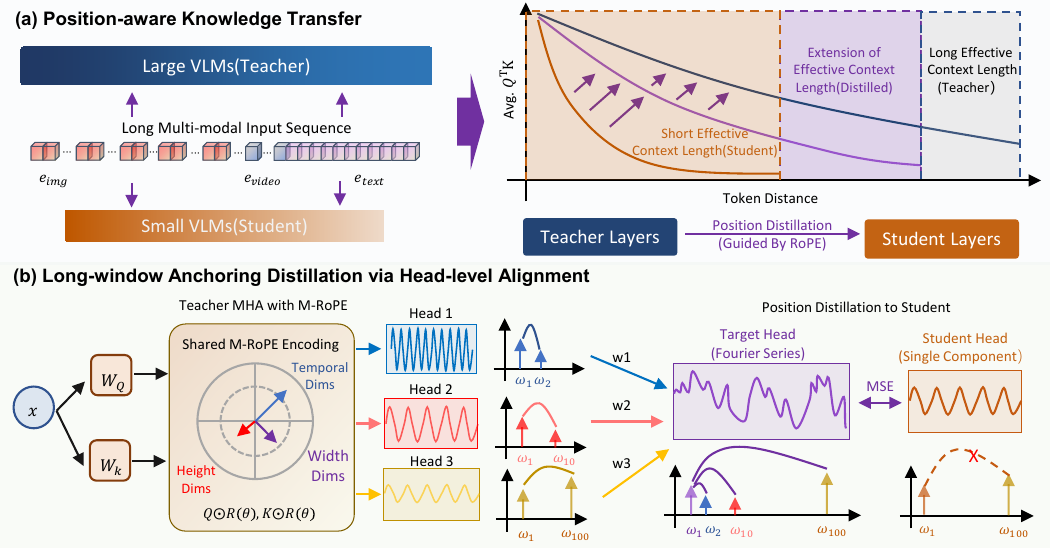}
    \caption{Overview of the LAid framework. \textbf{(a) Position-aware Knowledge Transfer:} LAid significantly extends the student's context length (purple) to approach the teacher's capability (gray), far exceeding the baseline (orange). \textbf{(b) Fourier-Enhanced Position Distillation:} Teacher attention heads capture positional information across frequency bands via mRoPE. We optimize weights ($w$) to distill these components into the student head, forming a rich Fourier series representation.}
    \label{fig:method}
\end{figure*}

\section{Related Works}

\textbf{Positional Encoding for Long Sequences.}
Rotary Position Embeddings (RoPE) and its variants (e.g., M-RoPE~\cite{qwen2-vl}, YaRN~\cite{YaRN}, HiRoPE~\cite{HiRoPE}, 3d-rpe~\cite{3d-rpe}, Liere~\cite{LieRE})  have become ubiquitous in modern VLMs due to their simplicity and effectiveness in capturing positional relationships.  A common strategy for context length extrapolation involves dynamically adjusting RoPE's rotational frequencies. Position Interpolation~\cite{PI} and YaRN~\cite{YaRN}apply frequency-based scaling to extend context windows from 4K to over 100K tokens, while LongRoPE~\cite{longRoPE} leverages non-uniform positional interpolation for multi-million token processing. Similarly, ABF~\cite{ABF} achieves 32K context by adjusting base frequency parameters with minimal fine-tuning. Alternative approaches restructure attention mechanisms: SelfExtend~\cite{SelfExtend} implements bi-level attention (grouped and neighbor attention) to efficiently handle long-range dependencies without fine-tuning, while DCA~\cite{DCA} decomposes sequences into manageable chunks with specialized attention patterns. MsPoE~\cite{MsPoE}and RandomPE~\cite{RandomPE}address the ``lost-in-the-middle'' problem through strategic position encoding modifications. PoSE~\cite{PoSE} simulates long inputs by manipulating position indices within short contexts. However, these approaches implicitly assume that positional knowledge transfers uniformly across model scales—overlooking critical variations in RoPE sensitivity between large and small VLMs. 

\textbf{Knowledge Distillation for VLMs.} 
Prior distillation efforts for VLMs have focused on task-specific performance (e.g., LLAVADI~\cite{llava-di} for VQA, DistillVLM~\cite{distillVLM} for instruction following) or cross-modal alignment (e.g., Align-KD~\cite{align-kd} for shallow-layer vision-text matching). While recent advances like VLM-KD~\cite{vlm-kd} address efficiency via text supervision or pruning, and methods such as LongReD~\cite{LongReD} explore distillation for long-context LLMs with RoPE, none explicitly optimize the long-context ability of VLMs.

\section{Methods}

 We formalize our task as follows: given a teacher VLM $\mathcal{T}$ with parameters $\theta_T$ (e.g., 32B) demonstrating effective context length $L_T$, and a student VLM $\mathcal{S}$ with $\theta_S$ (e.g., 7B) exhibiting $L_S < L_T$ despite architectural similarity, our goal is to extend $L_S \rightarrow L_S'$ where $L_S' \approx L_T$ through post-training alignment. We call it ``anchoring'' process, which must satisfy two constraints: (1) preserving $\mathcal{S}$'s computational efficiency during inference, and (2) maintaining performance on standard VL benchmarks. 


\subsection{Preliminary}


Given a large teacher VLM $\mathcal{T}$ with strong long-context modeling capabilities and a smaller student VLM $\mathcal{S}$, our objective is to enhance $\mathcal{S}$'s long-context performance through position-aware knowledge distillation. We first introduce the key concepts underlying our approach.

\textbf{Multi-head Attention.} The foundation of long-context ability in transformer architectures lies in the attention mechanism, which computes weighted interactions between sequence elements. In self-attention VLMs, these elements can be embedded as sub-word tokens or image patches. The standard attention operation is formulated as: $\text{Attention}(Q, K, V) = \text{softmax}\left({QK^T} /{\sqrt{d_k}}\right)V$, where $Q, K \in \mathbb{R}^{n \times d_k}$ and $V \in \mathbb{R}^{n \times d_v}$ represent the query, key, and value matrices, with $n$ being the sequence length and $d_k, d_v$ the embedding dimensions. Multi-head attention projects these matrices into multiple representation subspaces: $\text{MultiHead}(Q, K, V) = \text{Concat}(\text{head}_1, \ldots, \text{head}_h)W^O$, where $\text{head}_i = \text{Attention}(Q_i, K_i, V_i)$. This structure enables the model to jointly attend to information from different representation subspaces, capturing diverse dependency patterns.

\textbf{RoPE from a Fourier Perspective.} RoPE encodes positional information by applying a rotation matrix to query and key representations. For position $m$ and dimension $d$, the rotation operation is defined as: 
\begin{equation}
    \mathbf{q}_{m,d}^{\text{rot}} = R_{\theta}(m,d) \cdot \mathbf{q}_{m,d}, \
    \mathbf{k}_{m,d}^{\text{rot}} = R_{\theta}(m,d) \cdot \mathbf{k}_{m,d},
\end{equation}

where:
\begin{equation}
    R_{\theta}(m,d) = \begin{bmatrix} 
\cos(m\theta_d) & -\sin(m\theta_d) \\
\sin(m\theta_d) & \cos(m\theta_d)
\end{bmatrix},
\end{equation}

with $\theta_d = 10000^{-2d/D}$ and $D$ being the total embedding dimension. From a signal processing perspective, RoPE encodes positions using a spectrum of frequencies, forming a truncated Fourier series: $f_{\text{RoPE}}(m) = \sum_{d=0}^{D/2-1} \left[\cos(m\theta_d) + i\sin(m\theta_d)\right]$. As revealed by recent research~\cite{FoPE}, smaller models suffer more from \textbf{frequency leakage and distortion} when handling longer contexts, as their limited capacity constrains their ability to represent the full spectrum of necessary frequencies, leading to rapid attention decay over long distances.

\subsection{Head-level Position Alignment}

As illustrated in Figure~\ref{fig:method}(b), different attention heads in transformer models capture distinct aspects of contextual relationships. Research has shown that certain heads (position heads) in large models specialize in modeling long-range dependencies~\cite{position_head}, while others focus on local interactions. As model size increases, these position heads increasingly dominate the attention distribution, enabling superior long-context modeling.

To transfer this capability to smaller models, we propose a head-level alignment approach. For a student model layer $l$ with head index $i$ and teacher model layer $L$ with heads indexed by $j \in \{1,2,...,h_t\}$, we define our distillation objective as learning a set of weights $\{w_{i,j}\}$ such that:

\begin{equation}
    Q^s_{l,i} \approx \sum_{j=1}^{h_t} w_{i,j} \cdot Q^t_{L,j}, \quad
    K^s_{l,i} \approx \sum_{j=1}^{h_t} w_{i,j} \cdot K^t_{L,j},
\end{equation}

where $Q^s_{l,i}, K^s_{l,i}$ are the student's query and key matrices, and $Q^t_{L,j}, K^t_{L,j}$ are the teacher's counterparts. This allows each student head to learn from multiple teacher heads, with weights determining the contribution of each teacher head.

\subsection{Fourier Perspective on Position Distillation}

The core insight of our approach comes from viewing the distillation process through Fourier analysis. As shown in Figure 1(b), when RoPE is applied to the teacher model's query and key representations:

\begin{equation}
    Q^t_{L,j,\text{rot}} = Q^t_{L,j} \odot R_{\theta}(m), \quad
    K^t_{L,j,\text{rot}} = K^t_{L,j} \odot R_{\theta}(m).
\end{equation}

Our distillation process learns a linear combination of these frequency-encoded representations:

\begin{equation}
    Q^s_{l,i,\text{rot}} \approx \sum_{j=1}^{h_t} w_{i,j} \cdot (Q^t_{L,j} \odot R_{\theta}(m)).
\end{equation}

This can be interpreted as learning an enhanced rotational encoding:
\begin{equation}
    R'_{\theta}(m) = \sum_{j=1}^{h_t} w_{i,j} \cdot (W^Q_{t,j} \cdot R_{\theta}(m) \cdot (W^Q_{t,j})^{-1}),
\end{equation}

where $W^Q_{t,j}$ represents the teacher's query projection matrix for head $j$. This formulation enables the student to learn a richer Fourier series representation of positional relationships, expanding beyond the frequency limitations of standard RoPE and mitigating frequency leakage and distortion.

Figure~\ref{fig:method}(a) illustrates the resultant extension of effective context length achieved through our approach. The student model (purple curve) gains significantly enhanced long-range modeling capabilities compared to its original capacity (orange curve), approaching the performance of the much larger teacher model (gray curve).

Our complete distillation objective is formalized as:

\begin{equation}
\label{eq:fepd_loss}
\begin{split}
    \mathcal{L}_{\text{LAid}} = \sum_{l,i} \|Q^s_{l,i} - \sum_{j=1}^{h_t} w_{i,j} \cdot Q^t_{L,j}\|^2_F \\
    + \|K^s_{l,i} - \sum_{j=1}^{h_t} w_{i,j} \cdot K^t_{L,j}\|^2_F.
\end{split}
\end{equation}

This is combined with standard distillation losses:

\begin{equation}
    \mathcal{L}_{\text{total}} = \lambda_{\text{LAid}} \cdot \mathcal{L}_{\text{LAid}} + \lambda_{\text{KL}} \cdot \mathcal{L}_{\text{KL}} + \lambda_{\text{SFT}} \cdot \mathcal{L}_{\text{SFT}},
\end{equation}

where $\mathcal{L}_{\text{KL}} = \tau^2 \sum_{i} p^t_i(\tau) \log \frac{p^t_i(\tau)}{p^s_i(\tau)}$ is the KL-divergence between teacher and student output distributions with temperature \(\tau\), where \(p^t(\tau)\) and \(p^s(\tau)\) are the softened probability distributions of the teacher and student models, respectively, computed as \(p^t_i(\tau) = \frac{\exp(z^t_i / \tau)}{\sum_j \exp(z^t_j / \tau)}\) (similarly for \(p^s_i(\tau)\)), with \(z^t\) and \(z^s\) being the logits; and \(\mathcal{L}_{\text{SFT}}\) is a supervised fine-tuning loss.

\section{Experiments}
We address a practical context window extension scenario where only short-context training samples and limited computational resources are available. This constraint prevents straightforward gains from naive data length scaling, necessitating more efficient approaches that maximize the capabilities of student VLMs through architectural innovations or targeted training methods.

\begin{table*}[t]
\centering
\begin{tabular}{cccccc|ccc|cc}
\toprule
\multirow{2}{*}{\textbf{\#Params}} & \multirow{2}{*}{\textbf{Method}} & \multicolumn{4}{c|}{\textbf{Short Window \small{(\#img)}}} & \multicolumn{3}{c|}{\textbf{Long Window \small{(\#img)}}} & \multicolumn{2}{c}{\textbf{Avg. Gain ($\uparrow$)}} \\
\cmidrule(lr){3-6} \cmidrule(lr){7-9} \cmidrule(lr){10-11}
& & 1 & 5 & 10 & 20 & 50 & 100 & 150 & Short & Long \\
\midrule
32B & / & 83.79 & 79.11 & 74.71 & 73.34 & 68.17 & 62.56 & 60.65 & -- & -- \\
\midrule
\multirow{5}{*}{7B} & / & 80.22 & 68.45 & 62.19 & 57.21 & 54.73 & 51.08 & \underline{47.43} & -- & -- \\
& YaRN  & 80.03 & 63.78 & 62.09 & 55.96 & 56.26 & 47.96 & 42.36 & -2.5\% & -4.7\% \\
& SelfExtend & 78.53 & 62.58 & 58.69 & 53.01 & 50.35 & 45.12 & 40.12 & -5.9\% & -11.7\% \\
& SFT(LoRA) & \textbf{97.78} & \textbf{92.92} & \textbf{85.80} & \textbf{84.73} & \underline{63.10} & \underline{52.28} & 43.08 & \textbf{+35.92}\% & \underline{+3.6}\% \\
\rowcolor{gray!15} & \textbf{{\mn} (Ours)} & \underline{92.83} & \underline{83.26} & \underline{80.46} & \underline{74.09} & \textbf{67.04} & \textbf{63.37} & \textbf{60.17} & \underline{+24.1\%} & \textbf{+24.5\%} \\
\midrule
\multirow{5}{*}{3B} & / & 85.91 & 65.70 & 62.09 & 52.16 & 50.22 & 47.80 & 41.67 & -- & -- \\
& YaRN & 86.27 & 72.86 & 57.79 & 55.74 & 52.20 & 45.07 & 39.97 & +2.8\% & -1.9\% \\
& SelfExtend & 77.89 & 64.69 & 54.11 & 47.95 & 41.13 & 35.05 & 31.25 & -7.9\% & -23.26\% \\
& SFT(LoRA) & \textbf{98.20} & \textbf{91.88} & \textbf{87.48} & \textbf{68.89} & \underline{52.34} & \underline{48.01} & \underline{42.18} & \textbf{+31.5}\% & \underline{+1.96}\%\\
\rowcolor{gray!15} & \textbf{{\mn} (Ours)} & \underline{96.83} & \underline{83.34} & \underline{74.29} & \underline{63.27} & \textbf{58.2}& \textbf{53.91} & \textbf{50.23} & \underline{+20.1}\% & \textbf{+16.4}\% \\
\bottomrule
\end{tabular}
\begin{flushleft}
\small
Markers:
(1) \textbf{Bold} = the best performance per model size; 
(2) \textbf{Underline} = the second performance per model size;
(3) \cellcolor[rgb]{0.95,0.95,1.0}{{\mn}} = our proposed method;  
(4) ±\% = relative change compared to Base. 
(5) A vocabulary size mismatch between our 3B parameter model (151936) and the 32B teacher model (152064) renders the KL divergence loss $\mathcal{L}_\text{KL}$ incompatible. Consequently, the $\mathcal{L}_\text{KL}$ is excluded from the overall training objective $\mathcal{L}_\text{total}$.
Overall, higher accuracy indicates better performance.
(6) The rank of LoRA is 8
\end{flushleft}
\caption{Performance comparison of different context window extension methods for Vision-Language Models (VLMs) on Visual HayStack benchmark. Evaluated models include 32B/7B/3B parameter versions with five methods: Base (original), YaRN (RoPE extension), SelfExtend, SFT (supervised fine-tuning), and our {\mn}.}
\label{tab:main_results}
\end{table*}

\begin{table}[bt]
\centering
\label{tab:num_images_tokens}
\begin{tabular}{p{3cm} p{3cm}} 
\toprule
\textbf{\#Images} & \textbf{Avg. Tokens} \\
\midrule
1   & 393.46   \\
5   & 1849.58  \\
10  & 3630.64  \\
20  & 7401.08  \\
50  & 18237.85 \\
100 & 35418.29 \\
150 & 53653.22 \\
\bottomrule
\end{tabular}
\caption{Average input tokens by image count.}
\end{table}

\subsection{Experiments settings}

\textbf{Baselines.} Our comparison includes two categories of approaches: (1) \textit{Length Extrapolation} methods, including YaRN(Yet another RoPE extensioN method)~\cite{YaRN}, which employs frequency-based interpolation strategies for RoPE, and SelfExtend~\cite{SelfExtend}, which implements bi-level attention (grouped and neighbor attention) to capture both long-range and local dependencies. These methods enable context extension without requiring fine-tuning of pretrained LLMs. In VLMs, we extend them by applying YaRN's RoPE frequency interpolation to textual embeddings for longer text, and potentially to visual positional encodings. SelfExtend's bi-level attention is integrated via an additive mask—based on relative positions within combined visual-textual sequences—into the vision transformer and, critically, cross-modal attention layers, enabling efficient long-context multimodal interaction. (2) \textit{Supervised Fine-Tuning (SFT)}, where we directly fine-tune VLMs via LoRA~\cite{hu2022lora} on visual haystack tasks to enhance performance despite limited context.

\textbf{Models.} We evaluated our methodology using the Qwen2.5-VL models~\cite{qwen25vl}, a prominent family of open-source VLMs. Similar to the previous Qwen2-VL series~\cite{qwen2-vl}, Qwen2.5-VL incorporates Multimodal Rotary Position Embedding (M-RoPE) to effectively fuse positional information across textual, image, and video modalities. During their training phase, these models in 3B, 7B, 32B, and 72B parameter configurations, all support sequence lengths of up to 32,768 Specifically, we conducted a series of experiments utilizing the 7B parameter version as the student model and the 32B parameter version as its teacher model during the distillation progress.

\textbf{Datasets.} We adapt the Visual HayStacks (VHs) benchmark~\cite{visual_haystacks}, which is constructed from the COCO dataset with annotations at the object level~\cite{MSCOCO}. VHs presents binary (yes/no) questions about anchor and target objects in both single-needle (one relevant image) and multi-needle (multiple relevant images) settings. To create a practical evaluation environment, we construct a training set of 5,000 question-answer pairs with haystack sizes ranging from 2 to 20 images. For evaluation, we use 100 samples for each haystack size in the range [1, 2, 5, 10, 20, 50, 100, 150], enabling systematic assessment of model performance across increasing context lengths.

\begin{figure}[tb]
    \centering
    \includegraphics[width=\linewidth]{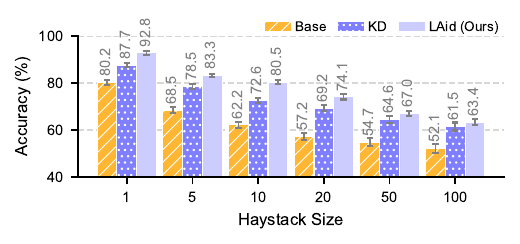}
    \caption{Distillation performance comparison on Visual HayStack dataset. The bar chart illustrates the accuracy of Base, Knowledge Distillation (KD), and our \mn~ method across increasing haystack sizes (1 to 100 images). Using Qwen2.5-VL-7B as the student model and Qwen2.5-VL-32B as the teacher, \mn~ consistently outperforms both baseline and standard KD approaches. 
    }
    \label{fig:FEPD_vs_KD}
\end{figure}

\begin{figure*}[btp]
    \centering
    \includegraphics[width=\linewidth]{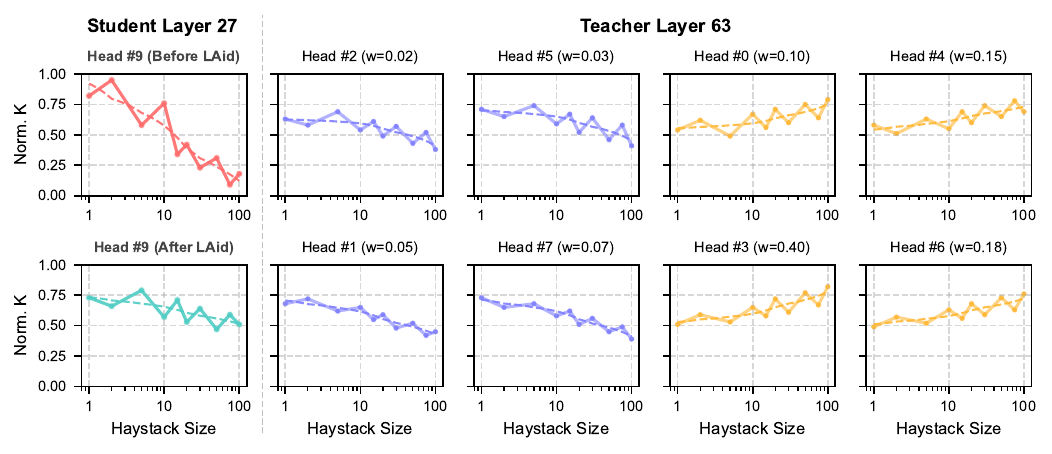}
    \caption{Head-level Knowledge Flow Analysis across different Visual HayStack Size. \textbf{The leftmost column} shows the student model's behavior, revealing a dramatic improvement from rapidly decaying activations (before \mn, top) to more stable patterns (after \mn, bottom). \textbf{The middle columns }display teachers' Local Position Heads (blue) with a slight downward trend across increasing context lengths, while\textbf{ the rightmost columns }show Global Position Heads (orange) that maintain or slightly increase activation values at longer distances. \mn~transfers this balanced position awareness to the student model, enabling it to maintain activation strength across the full range of context windows rather than focusing primarily on short-range dependencies.}
    \label{fig:head_knowledge_flow}
\end{figure*}

\label{exp:implement_details}
\textbf{Implementation Details.} We trained two student models, Qwen2.5-VL-7B-Instruct and Qwen2.5-VL-3B-Instruct, via knowledge distillation from the Qwen2.5-VL-32B-Instruct teacher model. Both student models were optimized using AdamW, with distinct learning rates set at $1 \times 10^{-5}$ for the student model parameters and $1 \times 10^{-4}$ for the weight coefficients. Training was conducted over 10 epochs on 4 NVIDIA A800 GPUs, employing an effective global batch size of 8, realized through a per-device batch size of 1 combined with 8 gradient accumulation steps. The maximum response sequence length was fixed at 512 tokens during training, and a learning rate warmup ratio of 0.05 was applied. We utilized an alpha value of 0.3 for knowledge distillation, specifically targeting the final layer (layer 27 for the 7B model and layer 35 for the 3B model). This training protocol resulted in total durations of approximately 74 hours for the 7B model and 43 hours for the 3B model.

\subsection{Main Results}

Our experimental evaluation reveals significant differences in how various window extension techniques perform on vision-language models, with results shown in Table \ref{tab:main_results}.

\textbf{Traditional Window Extension Methods Fail to Transfer to VLMs.} While large vision-language models (VLMs) demonstrate impressive long-context understanding, their smaller counterparts typically suffer from limited effective window lengths, restricting their utility in real-world applications. Interestingly, we observe that direct application of traditional context extension methods (YaRN, SelfExtend) yields suboptimal results on VLMs. YaRN shows a 4.7\% performance degradation on long contexts when applied to Qwen2.5-VL-7B, while SelfExtend performs even worse with an 11.7\% decline. This failure to transfer may stem from fundamental differences between unimodal and multimodal models. Unlike pure text LLMs, VLMs must maintain coherent cross-modal alignment across extended contexts, where visual positional embeddings interact with textual ones through complex attention patterns. Traditional methods optimize primarily for token-level dependencies without considering the unique spectral properties of multimodal attention, resulting in frequency distortion that particularly affects long-range visual-textual relationships.

\textbf{Supervised Fine-Tuning Exhibits Short-Context Bias.} Supervised fine-tuning (SFT) demonstrates remarkable performance improvements on short contexts (+35.92\% for Qwen2.5-VL-7B), but fails to maintain this advantage as context length increases, showing only modest gains (+3.6\%) on long contexts. This pattern reveals an inherent limitation of direct optimization approaches. The short-context bias of SFT can be attributed to its optimization objective, which prioritizes immediate performance gains without explicitly targeting the underlying mechanisms of long-range modeling. SFT effectively enhances content-based attention in familiar context ranges but fails to address the fundamental issue of positional attention decay at extended distances. This leads to overfitting on short-context patterns and poor generalization to longer sequences—precisely the opposite of what window anchoring aims to achieve.

\textbf{LAid Enables Balanced Long-Short Context Performance.} In this paper, we discover a surprising emergent property: knowledge distillation significantly enhances students' responsiveness to Rotary Position Embeddings (RoPE) at greater distances, directly correlating with extended context capabilities. Our proposed LAid method systematically exploits this phenomenon, achieving superior results across the full context spectrum. While LAid's improvements on short contexts (+24.1\%) are slightly below SFT's, it substantially outperforms all other methods on long contexts (+24.5\%), maintaining consistent performance even at extreme context lengths. This balanced capability stems from LAid's position-aware knowledge transfer approach, which preserves crucial spectral characteristics of the teacher model.

\begin{table*}[tb]
\centering
\begin{tabular}{cccccc|cc|cc}
\toprule
\multirow{2}{*}{\textbf{\#Params}} & \multirow{2}{*}{\textbf{Method}} & \multicolumn{4}{c|}{\textbf{Short Window \small{(\#img)}}} & \multicolumn{2}{c|}{\textbf{Long Window \small{(\#img)}}} & \multicolumn{2}{c}{\textbf{Avg. Gain ($\uparrow$)}} \\
\cmidrule(lr){3-6} \cmidrule(lr){7-8} \cmidrule(lr){9-10}
& & 1 & 5 & 10 & 20 & 50 & 100 & Short & Long \\
\midrule
\multirow{5}{*}{7B} & / & 80.22 & 68.45 & 62.19 & 57.21 & 54.73 & 47.43 & -- & -- \\
& w/o $\mathcal{L}_{\text{KL}}$ & \underline{91.26} & \underline{84.29} & \underline{75.09} & 65.97 & \underline{66.11} & \underline{62.29} & \underline{+18.2}\% & \underline{+20.2}\% \\
& w/o $\mathcal{L}_{\text{LAid}}$ & 87.68 & 78.50 & 72.57 & \underline{69.54} & 64.61 & 61.50 & +15.5\% & +18.1\% \\
\rowcolor{gray!15} & \textbf{{\mn} (Ours)} & \textbf{92.83} & \textbf{83.26} & \textbf{80.46} & \textbf{74.09} & \textbf{67.04} & \textbf{63.37} & \textbf{+24.1\%} & \textbf{+24.5\%} \\
\bottomrule
\end{tabular}
\begin{flushleft}
\small
Markers:
(1) \textbf{Bold} = the best performance per model size; 
(2) \textbf{Underline} = the second performance per model size;
(3) \cellcolor[rgb]{0.95,0.95,1.0}{{\mn}} = our proposed method;  
(4) ±\% = relative change compared to Base.
Overall, higher accuracy indicates better performance.
\end{flushleft}
\caption{Performance for loss ablations on VHs datasets, showing accuracy vs. number of images for each configuration.}
\label{tab:ablation}
\end{table*}

\subsection{Position-Aware Knowledge Distillation}

To dissect how position-aware distillation fundamentally differs from traditional approaches, we conduct a controlled comparison using Qwen2.5-VL-7B as student and Qwen2.5-VL-32B as teacher on the Visual HayStack benchmark. Figure \ref{fig:FEPD_vs_KD} presents performance across increasing context lengths (1 to 100 images), comparing our \mn~method against both the baseline model and standard knowledge distillation (KD). This decomposition allows us to isolate and quantify the specific contribution of position-aware distillation beyond conventional semantic knowledge transfer. The results reveal a clear pattern where \mn's advantage becomes increasingly pronounced at longer contexts, precisely where effective positional modeling is most critical.

\textbf{Performance Pattern.} Figure \ref{fig:FEPD_vs_KD} demonstrates \mn's consistent advantage over baseline and KD approaches. The performance gap is most significant at haystack size 1 (5.15\% over KD), and even with longer contexts, \mn~still achieves a notable 2.43\% improvement at size 50. While all methods show performance decline with increasing context length, \mn~maintains superior accuracy throughout the range.

\textbf{Distillation Limitations.} Traditional knowledge distillation transfers task-specific knowledge but fails to capture the position-sensitive representations essential for long-context modeling adequately. At haystack size 100, while KD improves over the baseline by 9.42\% (61.50\% vs. 52.08\%), it still falls short of \mn's 63.37\%. This confirms our hypothesis that conventional distillation approaches lack explicit mechanisms for transferring positional understanding.

\textbf{Spectral Enhancement.} \mn's superior performance stems from its Fourier-enhanced approach to position encoding. By transferring a richer Fourier series representation through our enhanced distillation objective (Eq. \ref{eq:fepd_loss}), \mn~enables smaller models to overcome the frequency leakage and distortion problems that typically plague them. This spectral enhancement is particularly evident in the consistent performance maintained across extended positional offsets.

\textbf{Balanced Context Modeling.} A distinctive advantage of \mn~is its simultaneous excellence in both short-context (1-20) and long-context (50-100) settings. Unlike other context extension methods that often sacrifice near-term accuracy for long-range performance, \mn~preserves performance across the entire context spectrum. This balanced capability derives from \mn's preservation of both high and low-frequency components in the positional encoding, enabling comprehensive modeling at all distance ranges.

\subsection{Head-level Knowledge Flow}

To elucidate the mechanisms of \mn, Figure \ref{fig:head_knowledge_flow} visualizes normalized key activation values across attention heads as context length grows. We analyze heads from Teacher Layer \#63 (Qwen2.5-VL-32B) and Student Layer \#27 (Qwen2.5-VL-7B), monitoring behavior over haystack sizes from 1 to 100 images. This reveals how \mn transfers positional awareness at the head level.

\textbf{Head Specialization.} The teacher shows distinct specialization: Local Position Heads (blue, middle) exhibit moderate activations (0.4-0.7) declining gradually with context, focusing on nearby tokens. Global Position Heads (orange, right) maintain stable or rising values (0.5-0.8) for long-range dependencies, evident in high-weight heads (e.g., Head \#3, w=0.4).

\textbf{Student Head Learning Pattern.} \mn~ transforms the student: Pre-distillation (top-left), activations decay rapidly from 0.8 to <0.2 due to frequency leakage. Post-\mn (bottom-left), they stabilize at 0.5-0.8, emulating a hybrid of teacher local/global behaviors.





\subsection{Ablation Study}

Our objective combines losses: 
\begin{equation}
\mathcal{L}_{\text{total}} = \lambda_{\text{LAid}} \cdot \mathcal{L}_{\text{LAid}} + \lambda_{\text{KL}} \cdot \mathcal{L}_{\text{KL}} + \lambda_{\text{SFT}} \cdot \mathcal{L}_{\text{SFT}}, 
\end{equation}

where \(\mathcal{L}_{\text{LAid}}\) aligns positional heads, \(\mathcal{L}_{\text{KL}}\) applies KL-divergence, and \(\mathcal{L}_{\text{SFT}}\) handles supervised fine-tuning.

We ablated components to evaluate contributions: 
\begin{enumerate}
    \item Full LAid Loss: Includes all components (\(\mathcal{L}_{\text{LAid}}\), \(\mathcal{L}_{\text{KL}}\), \(\mathcal{L}_{\text{SFT}}\)).
    
    \item LAid w/o $\mathcal{L}_{\text{LAid}}$: Trained with \(\mathcal{L}_{\text{LAid}}\) and \(\mathcal{L}_{\text{SFT}}\) only, excluding general knowledge distillation. 
    
    \item LAid w/o $\mathcal{L}_{\text{KL}}$: Trained with \(\mathcal{L}_{\text{LAid}}\) and \(\mathcal{L}_{\text{SFT}}\) only, excluding general knowledge distillation.
\end{enumerate}

Results in Table~\ref{tab:ablation} span 1-100 image contexts. Removing \(\mathcal{L}_{\text{LAid}}\) caused major drops (8.6\% short-context, 6.4\% long-context), confirming its role in long-range transfer via Fourier alignment. Ablating \(\mathcal{L}_{\text{KL}}\) yielded minor declines, emphasizing its support for general knowledge. The complete configuration achieved the best performance, demonstrating complementary effects among the loss functions.

In summary, \(\mathcal{L}_{\text{LAid}}\) drives context extension, with \(\mathcal{L}_{\text{KL}}\) aiding overall robustness; combined, they anchor smaller models to teacher long-window capabilities.

\section{Conclusions}

We present Long-window Anchoring distillation (LAid), tackling window-length gaps in VLMs. By enhancing RoPE responsiveness via position-aware distillation and Fourier-based head alignment, LAid preserves key low-frequency components overlooked by prior methods. Experiments show up to 3.2× context window extension, offering insights into positional transfer across scales.

Limitations include focus on attention (not feed-forward networks) and distillation overhead (unaffecting inference). Future work might integrate efficient tuning or combine with retrieval for ultra-long contexts.

\section{Acknowledgments}


This work was supported by the grants from the Natural Science Foundation of China (62225202, 62202029), Young Elite Scientists Sponsorship Program by CAST (No.2023QNRC001) and Beijing Natural Science Foundation (L248032). Thanks for the computing infrastructure provided by Beijing Advanced Innovation Center for Big Data and Brain Computing. This work is also sponsored by CAAI-MindSpore Open Fund, deve loped on OpenI Community. We owe sincere thanks to all authors for their valuable efforts and contributions. Jianxin Li is the corresponding author.

\bibliography{aaai2026}

\end{document}